\documentclass[journal]{IEEEtran}

\usepackage{graphics} 
\usepackage{epsfig} 
\usepackage{mathptmx}
\usepackage{times}
\usepackage{amsmath}
\usepackage{amssymb}
\usepackage{amsmath,amssymb}

\usepackage{graphicx}
\usepackage{subfigure}
\usepackage{caption}
\usepackage{mathrsfs}
\usepackage{algpseudocode}
\usepackage{algorithmicx, algorithm}
\usepackage{algpseudocode}
\usepackage{stfloats}
\usepackage{stfloats}

\begin{document}
%
\title{GAN-Based Interactive Reinforcement Learning from Demonstration and Human \\Evaluative Feedback}
\author{Jie Huang$^{1*}$, Rongshun Juan$^{1*}$, Randy Gomez$^{2}$, Keisuke Nakamura$^{2}$, Qixin Sha$^{1}$, Bo He$^{1}$, Guangliang Li$^{1**}$
	\thanks{$^{1}$College of Information Science and Engineering, Ocean University of China, 
		{\tt\small\{
			\{bhe, guangliangli\}@ouc.edu.cn}}
	\thanks{$^{2}$Honda Research Institute Japan Co., Ltd, Wako, Japan.
		{\tt\small\{r.gomez, keisuke\}@jp.honda-ri.com}}
	\thanks{$^{*}$ Contributing equally}
	\thanks{$^{**}$ Corresponding author}
}

\maketitle
\begin{abstract}
Deep reinforcement learning (DRL) has achieved great successes in many simulated tasks. The sample inefficiency problem makes applying traditional DRL methods to real-world robots a great challenge. 
Generative Adversarial Imitation Learning (GAIL) --- a general model-free imitation learning method, allows robots to directly learn policies from expert trajectories in large environments. However, GAIL shares the limitation of other imitation learning methods that they can seldom surpass the performance of demonstrations. In this paper, to address the limit of GAIL, we propose GAN-Based Interactive Reinforcement Learning (GAIRL) from demonstration and human evaluative feedback
by combining the advantages of GAIL and interactive reinforcement learning. We tested our proposed method in six physics-based control tasks, ranging from simple low-dimensional control tasks 
--- Cart Pole and Mountain Car, to difficult high-dimensional tasks --- Inverted Double Pendulum, Lunar Lander, Hopper and HalfCheetah. Our results suggest that 
with both optimal and suboptimal 
demonstrations, a GAIRL agent can always learn a more stable policy with optimal or close to optimal performance, while the performance of the GAIL agent is upper bounded by the performance of demonstrations or even worse than it. In addition, our results indicate the reason that GAIRL is superior over GAIL is the complementary effect of demonstrations and human evaluative feedback.
\end{abstract}

\IEEEpeerreviewmaketitle

\section{Introduction}
\IEEEPARstart{R}{einforcement} learning (RL) attempts to solve the challenge of building a robot that learns an optimal policy through interaction with the physical world via trial and error \cite{sutton1998reinforcement,kober2013reinforcement}. With recent advances in deep neural network, deep reinforcement learning (DRL) --- a combination of deep learning and RL, has achieved great successes in many simulated tasks, ranging from games \cite{mnih2015human,deepmind2019mastering,silver2017mastering} to complex locomotion behaviors \cite{heess2017emergence,florensa2017stochastic} and robotic manipulators \cite{openai2018learning}. However, shared by RL, DRL is sample inefficient and slow to converge because of the scarce reward signals \cite{goecks2020integrating}. It is difficult or even unpractical to design an efficient reward function for each task, which makes applying traditional DRL methods to real-world robots a great challenge \cite{riedmiller2018learning}.


Since most robots will operate in human inhabited environments, 
the ability to interact and learn from human users will be key to their success \cite{li2018interactive}. 
Many approaches for robot learning from interaction with a human user have been proposed. The interactive feedback that the human user provides during such interaction can take many forms, e.g., demonstrations, preferences, evaluative feedback, etc.
Among them, imitation learning from demonstrations was proposed and often leads to faster learning than learning from reward signals \cite{ross2010efficient,ng2000algorithms,abbeel2004apprenticeship,syed2008apprenticeship,syed2008game,ziebart2008maximum,bloem2014infinite}. In addition, it would be much easier to demonstrate behaviors than designing reward functions for robots to learn, though reward functions are commonly used to specify the objective of robots.

The simplest approach in this setting is behavioral cloning (BC) \cite{ross2010efficient}, in which the goal is to learn the mapping from states to optimal actions as a supervised learning problem. However, BC needs large amounts of data to learn and cannot generalize to unseen states, which is commonly used to initialize policies for RL \cite{rajeswaran2017learning,nagabandi2018neural}.
Another approach is inverse reinforcement learning (inverse RL), which learns a policy via RL, using a cost function extracted from expert trajectories \cite{ng2000algorithms}. 
Because of the assumption of expert optimality as prior on the space of policies, inverse RL can allow the learner to generalize expert behavior to unseen states more effectively \cite{ho2016model}. However, many of the proposed inverse RL algorithms need a model 
to solve a sequence of planning or reinforcement learning problems in an inner loop \cite{ho2016generative}, which is extremely expensive to run and limits their use in large and complex tasks. Moreover, their agents' performance might significantly degrade if the planning problems are not solved to optimality \cite{ho2016model,ermon2015learning}. Therefore, by drawing an analogy between imitation learning and generative adversarial networks (GANs) \cite{goodfellow2014generative}, Ho et al. proposed Generative Adversarial Imitation Learning (GAIL) --- a general model-free framework for directly learning policies from the expert trajectories \cite{ho2016generative}, and extended inverse RL to large environments. However, GAIL shares the limitation of the other imitation learning methods that they can seldom surpass the performance of demonstrations. 
In many tasks, demonstrations are not always optimal 
or the optimal demonstrations are hard to obtain practically. If the behavior of the demonstrator is suboptimal or far from optimal, the same will hold for GAIL. 
Moreover, 
the performance of policies trained through adversarial methods still falls short of those produced by manually designed reward functions, when such reward functions are available \cite{rajeswaran2017learning,peng2018variational,peng2018deepmimic}.

Fortunately, an agent via interactive reinforcement learning (interactive RL) 
from human evaluative feedback can generally surpass a trainer's performance in the task \cite{warnell2018deep,li2018interactive}. Based on reward shaping \cite{ng1999policy} in traditional RL, interactive RL from human evaluative feedback allows 
even non-technical people to train robots by evaluating their behaviors \cite{li2019human}. Hence, in this paper, we propose a model-free framework --- GAN-Based Interactive Reinforcement Learning 
(GAIRL), which combines the advantages of GAIL and interactive RL from human evaluative feedback. 
We hypothesize agents learn via GAIRL can outperform the demonstration performance and acquire an optimal policy disregarding the quality of demonstrations. 
We tested our method in six physics-based control tasks from the classic RL literature, ranging from simple low-dimensional control tasks 
--- Cart Pole and Mountain Car, to difficult high-dimensional tasks --- Inverted Double Pendulum, Lunar Lander, Hopper and HalfCheetah. Our results suggest that with both optimal and suboptimal demonstrations, a GAIRL agent can always learn a more stable policy with optimal or close to optimal performance, while the performance of the GAIL agent is upper bounded by the performance of demonstrations or even worse than it. In addition, our results indicate the reason that GAIRL is superior over GAIL might be because of the complementary effect of demonstrations and human evaluative feedback. 

\section{Related Work}

This section surveys the most related work to our approach in terms of imitation learning via inverse reinforcement learning and interactive reinforcement learning.

\subsection{Imitation Learning via Inverse Reinforcement Learning}

The inverse RL problem is to learn a policy via RL, using a cost function extracted from expert trajectories 
that prioritizes entire trajectories over others \cite{ng2000algorithms}. As a result, inverse RL does not suffer from compounding error problems as behavioral cloning \cite{ross2010efficient}. Moreover, because the assumption of expert optimality acts as prior on the space of policies, inverse RL can allow the learner to generalize expert behavior to unseen states more effectively \cite{ho2016model}.

In inverse RL, commonly, the reward function can be modeled via a linear combination of feature weights \cite{abbeel2004apprenticeship}. 
There are many inverse RL algorithms using this linear approximation for the reward function including apprenticeship learning 
\cite{abbeel2004apprenticeship}, maximum entropy inverse RL 
\cite{ziebart2008maximum}, etc. While many existing methods for apprenticeship learning output policies that are ``mixed'',  i.e., randomized combinations of stationary policies, apprenticeship learning using linear programming (LP) produces stationary policies \cite{syed2008apprenticeship}. However, it takes too much time for LP solver in an inner loop and the performance can significantly degrade if the planning problems are not solved to optimality \cite{ermon2015learning}. Based on a game-theoretic view of the problem, the game-theoretic 
apprenticeship learning is computationally faster, easier to implement, and even can be applied in the absence of an expert \cite{syed2008game}. However, recovering the agent's exact weights of linear approximation for the reward function is an ill-posed problem: multiple rewards can explain a single behavior \cite{ng2000algorithms}. Based on the principle of maximum entropy \cite{jaynes1957information}, D.Ziebart et al. \cite{ziebart2008maximum} proposed maximum entropy inverse RL 
to resolve the ambiguity in choosing a distribution over decisions. 


Most of above mentioned inverse RL methods need a model to 
solve a sequence of planning or reinforcement learning problems in an inner loop, which is extremely expensive to run and limit their use in large and complex tasks. Ho et al. \cite{ho2016model} proposed an imitation learning method by 
exploiting to learn a class of cost functions via distinguishing the expert policy from all others. Taking further inspiration from the success of nonlinear cost function classes in inverse RL \cite{ratliff2009learning}, Ho et al. \cite{ho2016generative} proposed GAIL by drawing an analogy between imitation learning and generative adversarial networks (GANs). GAIL is a general model-free framework for directly learning policies from the expert trajectories and extends inverse RL to large environments. 
Instead of learning from demonstrations supplied in the first-person, third-person imitation learning \cite{stadie2017third} improves upon GAIL by recovering a domain-agnostic representation of the agent's observations.
Generative adversarial imitation from observation \cite{torabi2018generative} learns directly from state-only demonstrations without having access to the demonstrator's actions by recovering the state-transition cost function of the expert.

However, GAIL inherits the problems of GANs, including possible training instability such as vanishing and exploding gradients, which stands out when the given expert demonstrations are not optimal \cite{arjovsky2017towards}.
By introducing a new type of variational autoencoder on demonstration trajectories, Wang et al. \cite{wang2017robust} increased the robustness of GAIL and avoided its mode collapse. 
In addition, Peng et al. \cite{peng2018variational} effectively modulated the discriminator's accuracy and maintain useful and informative gradients by enforcing a constraint on the mutual information between the observations and the discriminator's internal representation.
Instead of adversarial imitation learning, there are also 
many other 
methods 
integrating the preprocessed expert demonstrations into extensions of classic RL algorithms \cite{hester2018deep,gao2018reinforcement,reddy2019sqil,jing2020reinforcement}.
Our work 
overcomes the limitation 
shared by imitation learning that it seldom surpasses the performance of demonstrations by allowing it to learn from both demonstration and human evaluative feedback.


\subsection{Interactive Reinforcement Learning}

Inspired by potential-based reward shaping \cite{ng1999policy}, interactive RL is proposed to solve the sample efficiency problem in both RL and DRL, and allow an agent to learn from interaction with human at the same time. Specifically, 
an interactive RL agent can learn from non-experts in agent design and programming \cite{li2019human}. In interactive RL, a human trainer can evaluate the quality of an agent's behavior and give feedback to improve its 
behavior. For example, Thomaz and Breazeal \cite{thomaz2008teachable} implemented a tabular Q-learning \cite{watkins1992q} agent learning from environmental and human rewards. The TAMER agent learns from only human reward signal by directly modeling it \cite{knox2009interactively}. To facilitate an agent to learn in tasks with high-dimensional state space, Warnell et al. proposed deep TAMER using deep neural network to approximate the reward function \cite{warnell2018deep}. Loftin et al. \cite{loftin2016learning} take human feedback as as one kind of categorical feedback strategy for the agent to learn. The COACH algorithm learns by interpreting human reward signal as feedback to the current executing control policy of a robot \cite{macglashan2017interactive}. COACH was also extended to deep COACH using deep neural network as function approximator for the policy \cite{arumugam2019deep}. In addition, most related to our work, Li et al. proposed a method allowing an agent to learn from both human demonstration and evaluative feedback 
\cite{li2018interactive}. Our work differs by allowing an agent to learn in complex environment with a GAN-based model-free method.

\section{Background}

We consider an agent within the Markov decision process (MDP) framework. An MDP can be represented with a tuple $M = \{S,A,P,\mathbb{R},\gamma\}$. $\pi \in \Pi$ is a policy that takes an action $a \in A$ given a state $s \in S$. Successor states are derived from the dynamic model $P(s'\mid s,a)$. During the process, the agent will get a cost or reward $c(s,a)$ which is 
from a cost or reward function $C: S \times A \rightarrow \mathbb{R}$. $\overline{\mathbb{R}}$ denotes the extended real numbers $\mathbb{R} \bigcup \{+ \infty \}$. $\mathbb{E}_\pi[c(s,a)] \triangleq \mathbb{E}[\sum_{t=0}^T \gamma ^ tc(s_t,a_t)] $ denotes an expectation of the discounted return along the trajectory generated by policy $\pi$, where $\gamma$ is a discounted factor and $\gamma \in (0, 1]$. Similarly, $\mathbb{E}_\tau$ means an empirical expectation with respect to trajectory samples $\tau$. We use $\pi_E$ to refer to the expert policy and $\tau_E$ to refer to the expert trajectory samples.


\subsection{Maximum Entropy Inverse RL}

The first step of inverse RL 
is to learn a cost function based on the given 
expert demonstrations. The cost function is learned such that it is minimal for the trajectories demonstrated by the expert and maximal for every other policy \cite{abbeel2004apprenticeship}. Adding another constraint of choosing the policy with maximum entropy can 
solve the problem that many policies can lead to the same demonstrated trajectories. The general form of this framework can be expressed as:  
\begin{align}
IRL_\psi(\pi_E)=&\mathop{arg\max}\limits_{c\in\mathbb{R}^{S\times A}}-\psi(c)+(\mathop{min}\limits_{\pi\in\Pi}-H(\pi)+\nonumber\\
&\mathbb{E}_\pi[c(s,a)])-\mathbb{E}_{\pi_E}[c(s,a)]\label{con:coststep},
\end{align}
where $\psi(c):\mathbb{R}^{S\times A} \rightarrow \overline{\mathbb{R}}$ is a convex cost function regularizer, $H(\pi)$ is the entropy function of the policy $\pi$ defined by $H(\pi)\triangleq\mathbb{E}_\pi[-log \pi(a\mid s)]$. This step will output a desired cost function. Then the second step of maximum entropy inverse RL is an entropy-regularized RL step using the learned cost function, which can be defined as: 
\begin{equation}
RL(c)= \mathop{arg\min}\limits_{\pi\in\Pi}-H(\pi)+\mathbb{E}_\pi[c(s,a)]\label{con:rlstep}.
\end{equation}
The aim of the RL step is to find a policy minimizing the cost function and maxmizing the entropy.

\subsection{GAIL}
Ho et al. \cite{ho2016generative} replaced the $\psi(c)$ in Eq.\eqref{con:coststep} with a new cost regularizer $\psi_{GA}(c)$, which penalizes slightly on cost functions 
that 
assign small costs to expert state-action pairs and penalizes heavily otherwise. Instead of running RL after IRL, GAIL directly learns the optimal policy in the way that brings the distribution of the state-action pairs of the agent as close as possible to that of the demonstrator. The circuitous process summed up as Eq.\eqref{con:coststep} and Eq.\eqref{con:rlstep} can be solved by finding a saddle point ($\pi$, $D$) of the expression: 
\begin{equation}
- \lambda H(\pi) +\mathbb{E}_\pi[log(D(s,a))] + \mathbb{E}_{\pi_E}[log(1-D(s,a))]\label{con:gail},
\end{equation}
where a discriminator $D : S \times A \rightarrow (0,1)$ is trained to distinguish expert transitions $(s, a) \sim \tau_E$ from agent transitions $(s, a) \sim \tau_{agent}$ expressed as minimizing $\mathbb{E}_\pi[log(D(s, a))] + \mathbb{E}_{\pi_E}[log(1-D(s,a))]$. 
The agent is trained to ``fool" the discriminator into thinking itself as the expert expressed as maximizing $\mathbb{E}_\pi[log(D(s,a))]$. Taking $- \lambda H(\pi)$ out of Eq.\eqref{con:gail}, the loss function is analogous to that of GANs, which draws an analogy between imitation learning and GANs.  In our work, we use the discriminator network as a cost function for further RL step. Originally, the algorithm used to optimize the policy is Trust Region Policy Optimization (TRPO) \cite{schulman2015trust}. TRPO can prevents the policy from vibrating too much, as measured by KL divergence between the old policy and the new one averaged over the states in the trajectory samples, due to the noise in the policy gradient.

\subsection{Interactive RL from Human Evaluative Feedback}
Interactive RL was proposed to reduce the learning time of a RL agent based on reward shaping in traditional RL
\cite{ng1999policy}. Interactive shaping, one of the classical interactive RL methods, allows an agent to learn to perform a task by interpreting human evaluative feedback as reward signals as in traditional RL \cite{thomaz2008teachable, knox2009interactively}. In interactive RL from human evaluative feedback, a human trainer can observe the agent's behavior during the learning process and give evaluative feedback which will be used to update the agent's policy. 
That is to say, when an agent takes an action $a$ with respect to state $s$, the human trainer will provide 
evaluative feedback to imply the value of the selected action $a$. Then the agent will use this feedback as a reward to update the policy aiming to maximize cumulative rewards.

\section{Proposed Approach}
To address the limit of GAIL that it seldom surpasses the performance of demonstrations, in this paper, we propose a model-free framework --- GAN-Based Interactive Reinforcement Learning
(GAIRL) by 
combining the advantages of GAIL and interactive RL from human evaluative feedback. GAIRL is expected to 
leverage human demonstrations and human evaluative feedback to improve the training of agents and outperform the demonstrator disregarding the quality of the demonstrations. 

In GAIRL, besides the fixed cost function $c_{gail}(s,a)$ out of the discriminator as in GAIL, we 
introduce a human reward function into it 
as a new cost function, 
which can be expressed as: 
\begin{equation}
\label{cost_gairl}
c_{gairl}(s,a) = c_{gail}(s,a) + \alpha\mathbb{H}(s,a),
\end{equation}
where $\mathbb{H}(s,a)$ is the human reward function approximated with a neural network (HRN), and $\alpha$ is a weight vector for balancing the cost function $c_{gail}(s,a)$ out of the discriminator in GAIL and the learned HRN. Specifically, we set 
\begin{equation}
\label{cost}
c_{gail}(s,a) = -lg(\frac{e^{-D(s,a)}}{1+e^{-D(s,a)}}+10^{-8}).
\end{equation} 
In addition, Ho et al. \cite{ho2016generative} defined a policy's occupancy measure $\rho_\pi : S \times A \rightarrow \mathbb{R}$ as 
\begin{equation}
\rho_\pi(s,a) = \pi(a{\mid}s)\sum_{t=0}^\infty\gamma^tP(s_t=s{\mid}\pi),
\end{equation}
which means the unnormalized distribution of state-action pairs. They proved that 
\begin{equation}
RL \circ IRL_{\psi_{GA}}(\pi_E) = arg\min_{\pi\in\Pi} -\lambda H(\pi) + \psi_{GA}^\ast(\rho_\pi - \rho_{\pi_E}),
\end{equation}
where $\psi_{GA}^\ast$ is the convex conjugate of convex function $\psi_{GA}$. So we can try to solve $arg\min_{\pi\in\Pi} -\lambda H(\pi) + \psi_{GA}^\ast(\rho_\pi - \rho_{\pi_E})$ directly instead of running RL after IRL as in other inverse RL methods. 
In our method, we use a new cost regularizer derived from $\psi_{GA}$, which can be expressed as
\begin{equation}
\psi_{gairl}  \triangleq \left\{
\begin{array}{rcl}
&\mathbb{E}_{\pi_E}[\mbox{g}(c_{gairl}(s,a))] &   {if{\quad}c_{gairl}< 0}\\
&+\infty                                      &         {otherwise},
\end{array}\right.
\end{equation}
where 
\begin{equation}
g(x) = \left\{
\begin{array}{rcl}
&-x - log(1 - e^x) &   {if{\quad}x<0}\\
&+\infty           &     {otherwise}.
\end{array}\right.
\end{equation}
Its convex conjugate is shown as
\begin{align}
\psi_{gairl}^\ast(\rho_\pi - \rho_{\pi_E}) = &\mathop{\max}\limits_{D\in(0,1)^{S\times A}} \sum_{s,a}\rho_\pi(s,a)log(D(s,a))+\nonumber\\
&\rho_{\pi_E}(s,a)log(1-D(s,a)),
\end{align}
where $D:S\times A\rightarrow (0,1)$ is a discriminative classifier.  Finally, our proposed approach can be summarized as solving
\begin{align}
\mathop{\min}\limits_{\pi\in\Pi} \psi_{gairl}^\ast(\rho_\pi - \rho_{\pi_E}) = &\mathop{\min}\limits_{\pi\in\Pi} \mathop{\max}\limits_{D\in(0,1)^{S\times A}} \mathbb{E}_\pi[log(D(s,a))]+\nonumber\\&\mathbb{E}_{\pi_E}[log(1-D(s,a))] -\lambda H(\pi)
\end{align}
and then performing a RL step expressed as
\begin{equation}
RL(c_{gairl})= \mathop{arg\min}\limits_{\pi\in\Pi}\mathbb{E}_\pi[c_{gairl}(s,a)].
\end{equation}

\renewcommand{\algorithmicrequire}{\bf{Input:}}
\renewcommand{\algorithmicensure}{\bf{Outputput:}}
\begin{algorithm}[t]
	\setcounter{algorithm}{0}
	\caption{Part 1}
	\begin{algorithmic}[1]
		\Require
		Demonstrations $\mathcal{E}$, initialize agent's policy $\pi_0$, discriminator $D_0$ and HRN $\mathbb{H}$, Replay Buffer $\mathcal{B}_1$, $\mathcal{B}_2$. 
		\Ensure  
		Trained discriminator $D$ and HRN $\mathbb{H}$, agent's policy $\pi_i$
		\State Assign a human reward of $R_h = +1$ to all the state-action pairs $[s_d,a_d]$ from $\mathcal{E}$
		\State Store all the tuples $[s_d,a_d,+1]$ in $\mathcal{B}_1$, $\mathcal{B}_2$ respectively
		\State $i = 0$
		\Repeat
		\State Sample agent trajectories $\tau_i$ from $\pi_i$ and demonstrated 
		\Statex \quad\ \ trajectories $\tau_E$ from $\mathcal{E}$ 
		\If {Receive human feedback $R_h$ with respect to $[s, a]$ \qquad \qquad from $\tau_i$} store $[s, a, R_h]$ in $\mathcal{B}_1$
		\State Randomly sample a batch of $[s,a, R_h]$ from $\mathcal{B}_1$
		\State Update $\mathbb{H}$ using Rmsprop with loss $\mathcal{L}_{hrn}$
		\EndIf
		\State {Update $D_i$ using Adam with loss  \Statex \qquad\quad$-(\mathbb{E}_{\tau_i}[log(D_i(s,a))]+\mathbb{E}_{\tau_E}[log(1-D_i(s,a))])$}
		\State {Update $\pi_i$ using TRPO with loss \Statex \qquad\qquad\qquad $-\lambda H(\pi)+\mathbb{E}_{\tau_i}[log(D_{i+1}(s,a))] $ }
		\State $i = i + 1$
		\Until {Meet Discriminator Requirements}
		\State $D = D_i$
		\algstore{bkbreak}
	\end{algorithmic}
\end{algorithm}
The proposed method is summarized in Algorithm 1, 
which contains two main steps. The first main step is the GAIL alike step, as shown in Algorithm 1 Part 1. Similar to GAIL, this step starts by sampling demonstrated trajectories from demonstrations and agent trajectories from the agent's current policy to perform the update on the discriminator. 
However, different from GAIL, 
in GAIRL, a human trainer, who might be non-expert in agent design or even in programming, 
can provide evaluative feedback by evaluating the agent's behavior according to her knowledge in the task. 
We use human evaluative feedback as labels of corresponding samples to train a human reward network (HRN) for predicting it. 
The standard mean square error is used as the loss function of HRN. Specifically, when the agent takes an action $a$ in state $s$, a human trainer gives a reward $R_h$ based on the evaluation of the state and action. Then we store the tuple $[s,a,R_h]$ in the replay buffer for HRN and randomly sample the tuple $[s,a,R_h]$ from the replay buffer to update HRN by 
minimizing the loss:
\begin{equation}
\mathcal{L}_{hrn} =\frac{1}{n}\sum_{i=0}^n(R_h-\mathbb{H}(s,a)),
\end{equation}
where $\mathbb{H}(s,a)$ is the estimated HRN. The human reward $R_h$ is defined as below:
\begin{equation}
R_h = \left\{
\begin{array}{rcl}
+N&  &agent\ reaches\ the\ goal\\
+1&  &good\ action\\
-1&  &bad\ action\\
-N&  &agent\ fails,
\end{array}\right.
\end{equation}
where $N$ is determined by different tasks (2 for Cart Pole, 6 for Mountain Car, 4 for Inverted Double Pendulum, 5 for Lunar Lander, Hopper and HalfCheetah). 
In addition, to ease the burden of the human trainer, we assign a human reward of $R_h = +1$ to all 
the state-action pairs $[s_d, a_d] $ in the demonstrations, 
and store all the tuples $[s_d, a_d, +1]$ in the replay buffer for HRN 
to obtain sufficient samples with ``good action''.

Both the HRN and the discriminator network will be used in the local cost function in Eq. (\ref{cost_gairl}) for providing learning signals to the policy and perform the update on it with TRPO. However, 
in this step, $\alpha$ is set to be a small value, so that we can update the discriminator network while learning the human reward network simultaneously with little effect on the cost function. That is to say, the human reward function has little 
effect on the policy learning with TRPO. 
Repeating the above steps until both the discriminator's expert accuracy and the generator accuracy reach 0.99, which can output a good, fixed cost function to effectively distinguish expert's state-action pairs from the agent's state-action pairs.   
\begin{algorithm}[h]
	\setcounter{algorithm}{0}
	\caption{Part 2}
	\begin {algorithmic}[1]
	\algrestore{bkbreak}
	\While {Policy Improves}
	\State Sample agent trajectories $\tau_i$ from $\pi_i$
	\If {Receive human feedback $R_h$ with respect to $[s, a]$ \qquad \qquad from $\tau_i$} store $[s, a, R_h]$ in $\mathcal{B}_2$
	\State Randomly sample a batch of $[s,a, R_h]$ from $\mathcal{B}_2$
	\State Update $\mathbb{H}$ using Rmsprop with loss $\mathcal{L}_{hrn}$
	\EndIf
	\State Update $\pi_i$ using DQN/TD3 with reward
	\Statex \quad\  \qquad $c_{gairl}(s,a) = c_{gail}(s,a) + \alpha\mathbb{H}(s,a)$ 
	\State $i = i + 1$
	\EndWhile
\end{algorithmic}
\end{algorithm}

The second main step in GAIRL is the RL step, as shown in Algorithm 1 Part 2. By setting $\alpha$ to be a large value, the new cost function $c_{gairl}(s,a)$ consisting of both the cost function $c_{gail}(s,a)$ from the discriminator in GAIL and learned HRN will serve as a reward function for further RL method. In addition, the human trainer can further provide evaluative feedback to train HRN. In this case, both the cost function $c_{gail}(s,a)$ learned from demonstrations and estimated HRN will provide rewards for the agent to perform updates on the policy. 
In our experiment, the value of $\alpha$ in the second main step varies in 
different tasks and will be experimentally optimized 
to achieve the best performance for a GAIRL agent. 
In our work, we choose DQN \cite{mnih2013playing} for the RL step in tasks with continuous observation space and discrete action space  
and Twin Delayed Deep Deterministic policy gradient (TD3) \cite{fujimoto2018addressing} in tasks with continuous observation and continuous action space. 

\section{Experiments}
\begin{figure}[thpb]
	\vspace{-4mm}
	\centering
	\subfigure[Cart Pole] {
		\begin{minipage}[t]{0.4\linewidth}
			\centering
			\includegraphics[width=1in,height=0.6in]{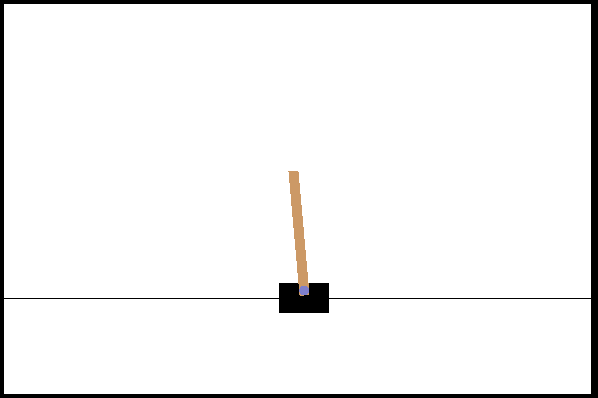}
	\end{minipage}}%
	\subfigure[Mountain Car]{
		\begin{minipage}[t]{0.4\linewidth}
			\centering
			\includegraphics[width=1in,height=0.6in]{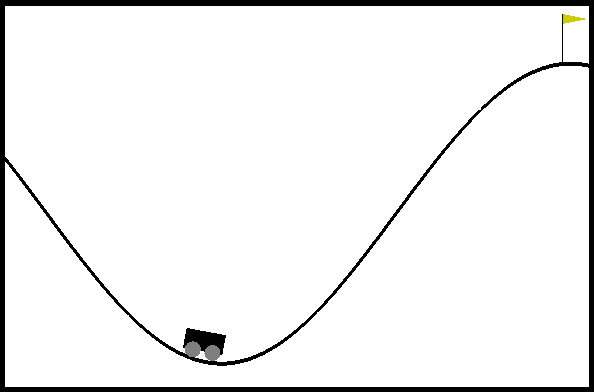}
	\end{minipage}}
	\subfigure[Inverted Double Pendulum]{
		\begin{minipage}[t]{0.4\linewidth}
			\centering
			\includegraphics[width=1in,height=0.6in]{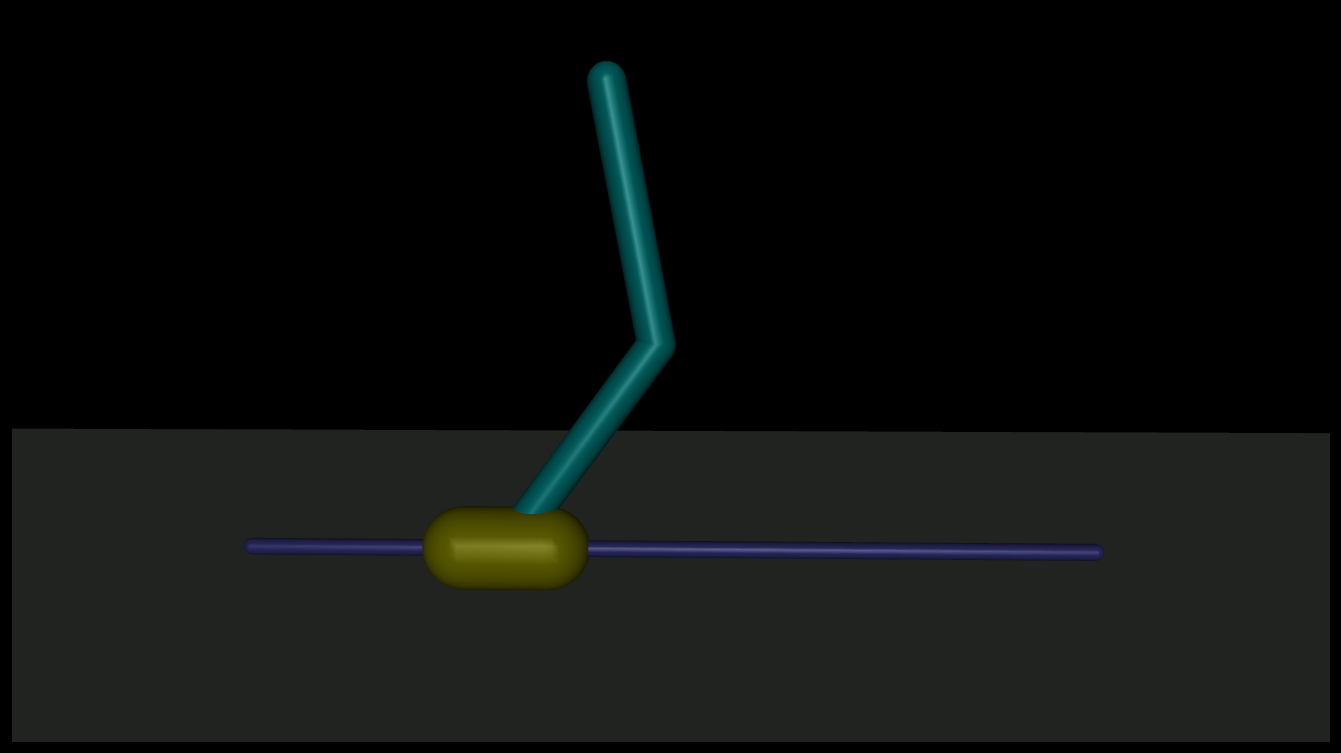}
	\end{minipage}}
	\subfigure[Lunar Lander]{
		\begin{minipage}[t]{0.4\linewidth}
			\centering
			\includegraphics[width=1in,height=0.6in]{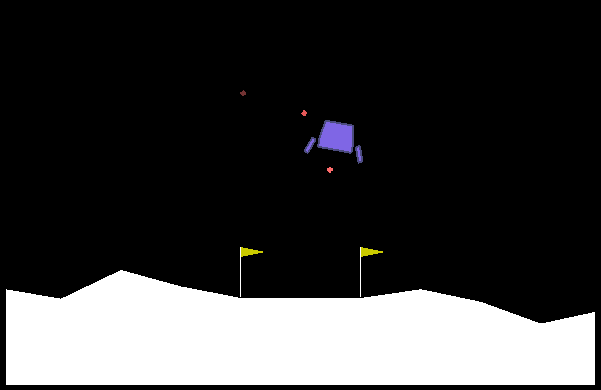}
	\end{minipage}}
	\subfigure[Hopper]{
		\begin{minipage}[t]{0.4\linewidth}
			\centering
			\includegraphics[width=1in,height=0.6in]{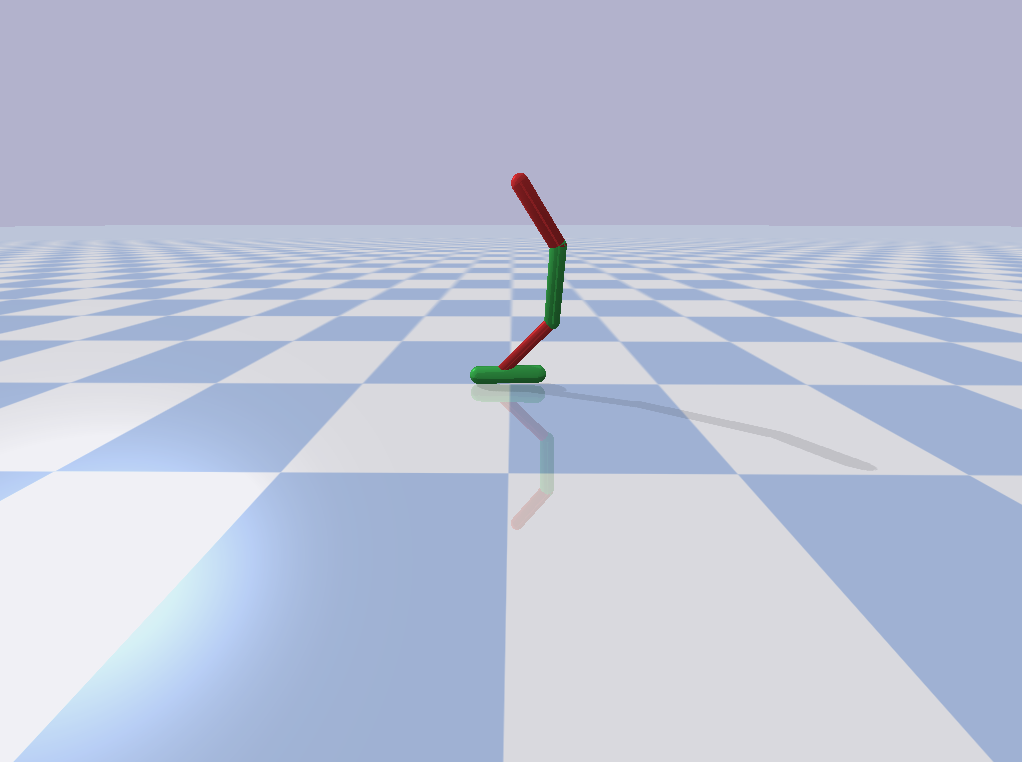}
	\end{minipage}}
	\subfigure[HalfCheetah]{
		\begin{minipage}[t]{0.4\linewidth}
			\centering
			\includegraphics[width=1in,height=0.6in]{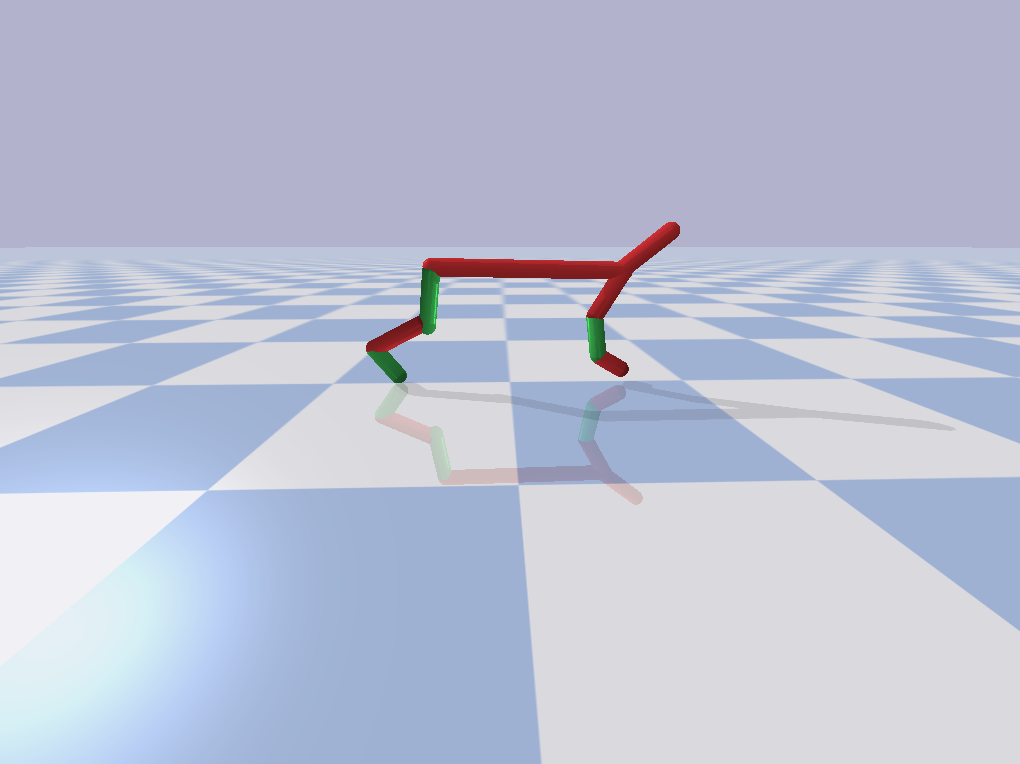}
	\end{minipage}}
	\caption{Screenshots of the six tasks.} 
	\label{tasks}
	\vspace{-6mm}
\end{figure}

We tested our proposed method --- GAIRL --- in six physics-based control tasks, ranging from low-dimensional to high-dimensional tasks from the classic RL literature:  Cart Pole, Mountain Car, Inverted Double Pendulum, Lunar Lander, Hopper and HalfCheetah. 
The Inverted Double Pendulum task is simulated with MuJoCo \cite{todorov2012mujoco}, Hopper and HalfCheetah are simulated with Pybullet \cite{coumans2016pybullet}, while other tasks are from OpenAI Gym \cite{brockman2016openai}.  
Figure \ref{tasks} shows screenshots of the six tasks.

\subsection{Experimental Tasks}
\label{tasksinfo}

$\bf{Cart}$ $\bf{Pole}$: A pole is attached by an un-actuated joint to a cart, which moves along a frictionless track. The pole starts upright, and the goal is to prevent it from falling over by increasing and reducing the cart's velocity. The state space is 
four-dimensional continuous 
and the action space is 
two-dimensional discrete. 
The reward is 1 for every step taken.


$\bf{Mountain}$ $\bf{Car}$: A car started from the bottom of a valley and its goal is to climb to the top of the mountain. The state space is two-dimensional continuous 
and action space is 
three-dimensional discrete. 
The reward is given based on the vertical distance from the agent's position to the bottom of a valley, and a custom reward function $\mit{\left|s[0] - (-0.06)\right|}$ instead of the reward function defined in Gym is used. 

$\bf{Inverted}$ $\bf{Double}$ $\bf{Pendulum}$: This is a complex version of Cart Pole in 3D environment. 
The 2-link pendulum starts upright, and the goal is to prevent it from falling over by increasing and reducing the sliding block's velocity. The state space is 
eleven-dimensional continuous 
and the action space is 
one-dimensional continuous. 
The reward is given at every step based on the angle of the pendulum at the end.

$\bf{Lunar}$ $\bf{Lander}$: This is a simulation environment for testing and solving the rocket trajectory optimization --- a classical  optimal control problem. The state space is 
eight-dimensional continuous 
and the action space is 
four-dimensional discrete. 
The reward is given at every step based on the relative motion of the lander with respect to the landing pad. 

$\bf{Hopper} \& \bf{HalfCheetah} $: The goal of these tasks are to let the hopper and half part of a cheetah go forward as far as possible. In Hopper, the state space is fifteen-dimensional continuous and the action space is three-dimensional continuous. In HalfCheetah, the state space is 26-dimensional continuous and the action space is six-dimensional continuous. Reward is given at every step based on their relative motions with respect to the floor, control cost, etc. 

Note that all these rewards mentioned above are only used for evaluating agents' performance and generating demonstrations, but never for learning.

\subsection{Experimental Setup} 
In each task, we train five agents: a GAIL agent as baseline; a GAIRL agent 
learning from rewards provided by the cost function $c_{gairl}(s,a)$ extracted from demonstrations and human reward function learned during the GAIL alike step; a DQND or TD3D agent learning from 
the cost function $c_{gail}(s,a)$ extracted from expert demonstrations  
during the GAIL alike step;  a DQNH or TD3H agent learning from 
human reward function learned during the GAIL alike step; a BC agent running imitation learning by behavioral cloning. For each task, 
a true reward function from OpenAI Gym \cite{brockman2016openai} as described in Section \ref{tasksinfo} 
is used for evaluating the agent performance, but never 
for learning. 

During experiments, we first generate demonstrated behavior for these tasks by running DQN/TD3 on these true reward functions to create demonstrated policies of different qualities --- optimal or suboptimal, instead of using human experts to provide demonstrations. 
Then we sample datasets of trajectories 
from the demonstrated policies, each consisting of about 200 state-action pairs. For all experiments, 
we use 10 demonstrations, 
as suggested by Ho et al. \cite{ho2016generative}. The first authors trained the human reward network HRN for all tasks. The weights of entropy term $\lambda$ in our work is $10^{-3}$.

We trained all agents' policies 
of the same neural network architecture for all tasks: two hidden layers of 100 units each, with $tanh$ nonlinearities in between. The architecture of discriminator network in GAIRL is the same as GAIL. 
The human reward network has two hidden layers of 100 units each, with $Relu$ nonlinearities in between and $tanh$ nonlinearities multiplying $N$ in the output layer. All networks were always initialized randomly at the start of each trial. 
In each task, three random seeds are used for the environment simulator and the network initialization.
Hopper and HalfCheetah tasks are run for 2 million timesteps, while the other tasks are run for 1 million timesteps. The performance metric, if not specified, used throughout the experiments is the mean accumulated reward per 100 episodes in terms of the true reward function from Gym. 

\section{Results and Discussion}

\begin{figure*}[ht]
	\vspace{-5mm}
	\centering
	\includegraphics[scale=0.34]{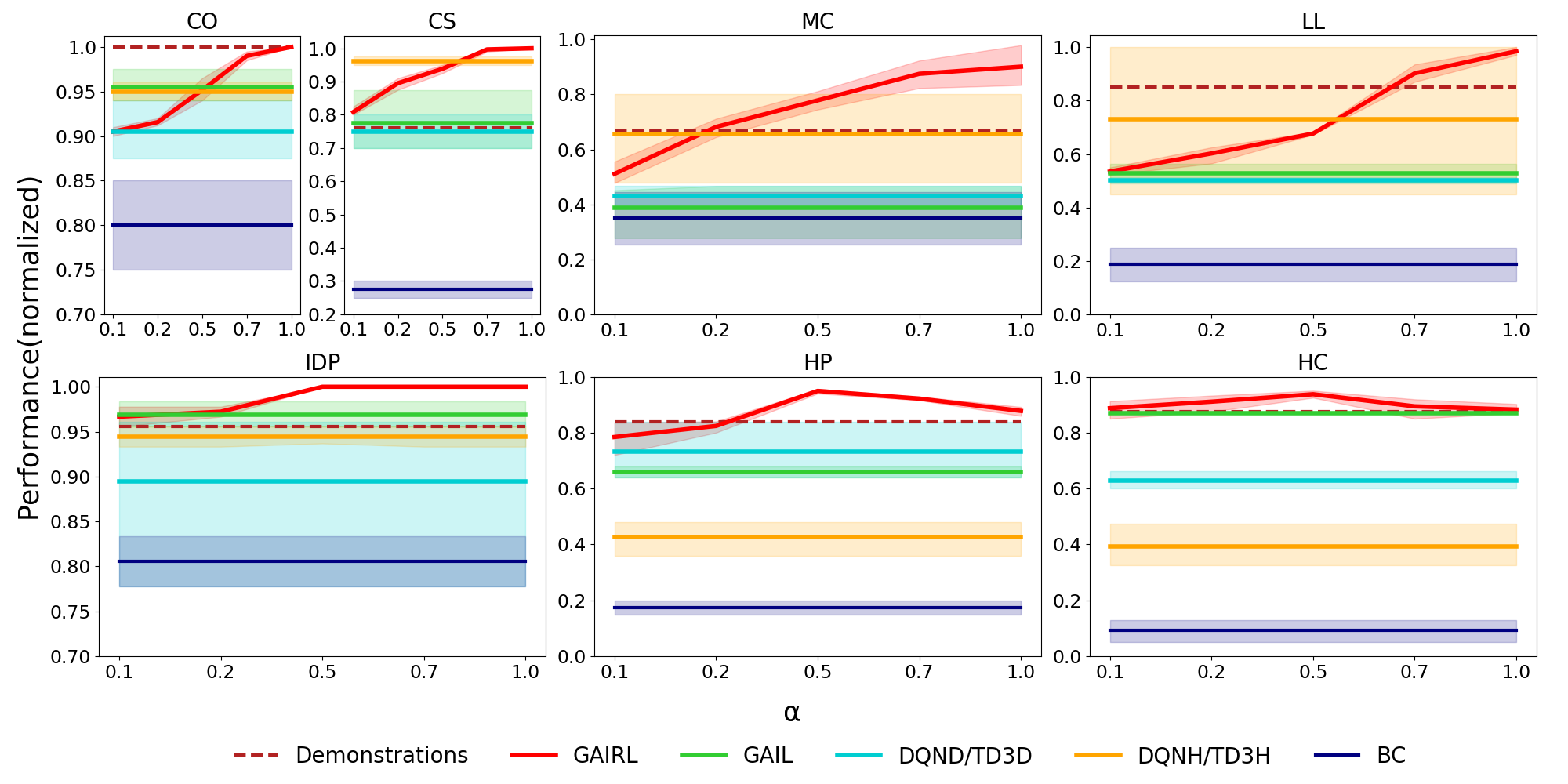}
	\caption{Effect of 
	the learned HRN on the final performance of the five agents. 
	The y-axis is the normalized mean accumulated reward per 100 episodes with optimal performance as 1 in different tasks. 
	The x-axis is $\alpha$ for balancing the cost function out of the discriminator in GAIL and 
	HRN. Note: CO---Cart Pole with Optimal Demonstration, CS---Cart Pole with Suboptimal Demonstration, MC--- Mountain Car, 
	LL---Lunar Lander, 
	IDP---Inverted Double Pendulum, 
	HP---Hopper, 
	HC---HalfCheetah.} 
	\label{FP}
	\vspace{-3mm}
\end{figure*}

In this section, we present and analyze the experimental results 
by comparing the performances of the five trained agents: GAIL agent, GAIRL agent, DQND/TD3D agent, DQNH/TD3H agent, BC agent. 
The shaded area is the 0.95 confidence interval and the bold line is the mean performance. Note that optimal performance and the performance of demonstrations with a confidence interval are also shown for reference.
Both optimal and suboptimal demonstrations are offered in the Cart Pole task; only suboptimal demonstrations are offered in other tasks.

\subsection{Effect of HRN on Agent's Performance} 

%

We first analyze the effect of human evaluative feedback on agent's learning by comparing the five agents' final performance with varied $\alpha$ which is a weight vector for balancing the cost function $c_{gail}(s,a)$ out of the discriminator in GAIL and the learned HRN from human evaluative feedback, as in Eq. (\ref{cost_gairl}). Figure \ref{FP} shows the normalized final performance of the five agents in the six tasks. From Figure \ref{FP} we can see that, in relatively simple tasks (e.g., Cart Pole, Mountain Car, Lunar Lander), as $\alpha$ increases, the final performance of GAIRL agent goes up to optimal with optimal demonstrations and close to optimal even with suboptimal demonstrations. In Inverted Double Pendulum, Hopper and HalfCheetah, the final performance of GAIRL agent only improves and reaches close to optimal when $\alpha$ increases from 0.1 to 0.5, then stagnates in Inverted Double Pendulum and decreases in Hopper and HalfCheetah. Our results show that with a proper value of $\alpha$ in all tasks, the GAIRL agent can learn much better than learning from demonstrations (GAIL, DQND/TD3D, BC) and human evaluative feedback (DQNH/TD3H) alone separately, achieving optimal or close to optimal performance.


\subsection{Learning Curve} 
We further analyze the effect of human evaluative feedback on GAIRL agent's performance in the learning process by comparing the learning curves of the four trained agents: GAIL agent, GAIRL agent, DQND/TD3D agent, DQNH/TD3H agent. BC agent is not shown because it does not learn interactively. Note that $\alpha$ is set to be able to achieve the best performance in all tasks (1 for Cart Pole, Mountain Car and Lunar Lander, 0.5 for Inverted Double Pendulum, Hopper and HalfCheetah). 

\subsubsection{Low-dimensional Tasks}
\begin{figure}[thpb]
	\vspace{-1mm}
	\centering
	\includegraphics[width=3.6in,height=2in]{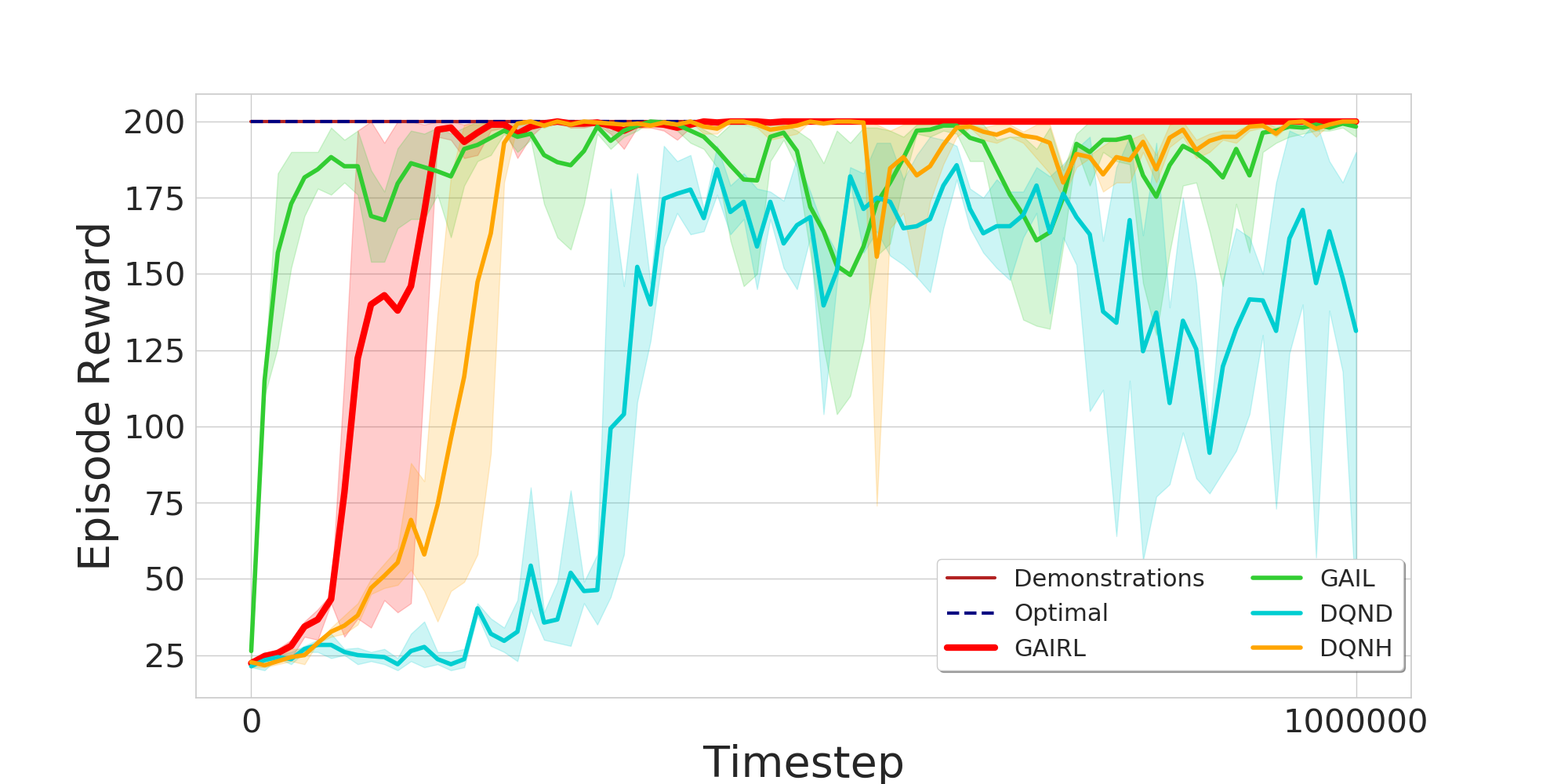}
	\caption{The four agents' learning curves with optimal demonstrations in Cart Pole.} 
	\label{cartpoleO}
	\vspace{-3mm}
\end{figure}

Figure \ref{cartpoleO} shows four agents' learning curves with optimal demonstrations in Cart Pole. From Figure \ref{cartpoleO} we can see that, with optimal demonstrations, the performance of GAIL agent almost immediately reaches the first peak which is faster but worse than the GAIRL agent. 
However, the performance of the GAIRL agent in the first peak is already close to optimal and stabilizes shortly after that, 
while the performance of GAIL agent still fluctuates below the optimal performance with an unstable policy reaching the optimal performance occasionally. 
Both DQND and DQNH agents learn 
slower than the GAIL and GAIRL agent. The performance of the DQND agent fluctuates in a similar way as the GAIL agent, while the DQNH agent learns faster than the DQND agent and reaches the optimal performance after final training. 

\begin{figure}[thpb]
	\centering
	\vspace{-3mm}
	\includegraphics[width=3.6in,height=2in]{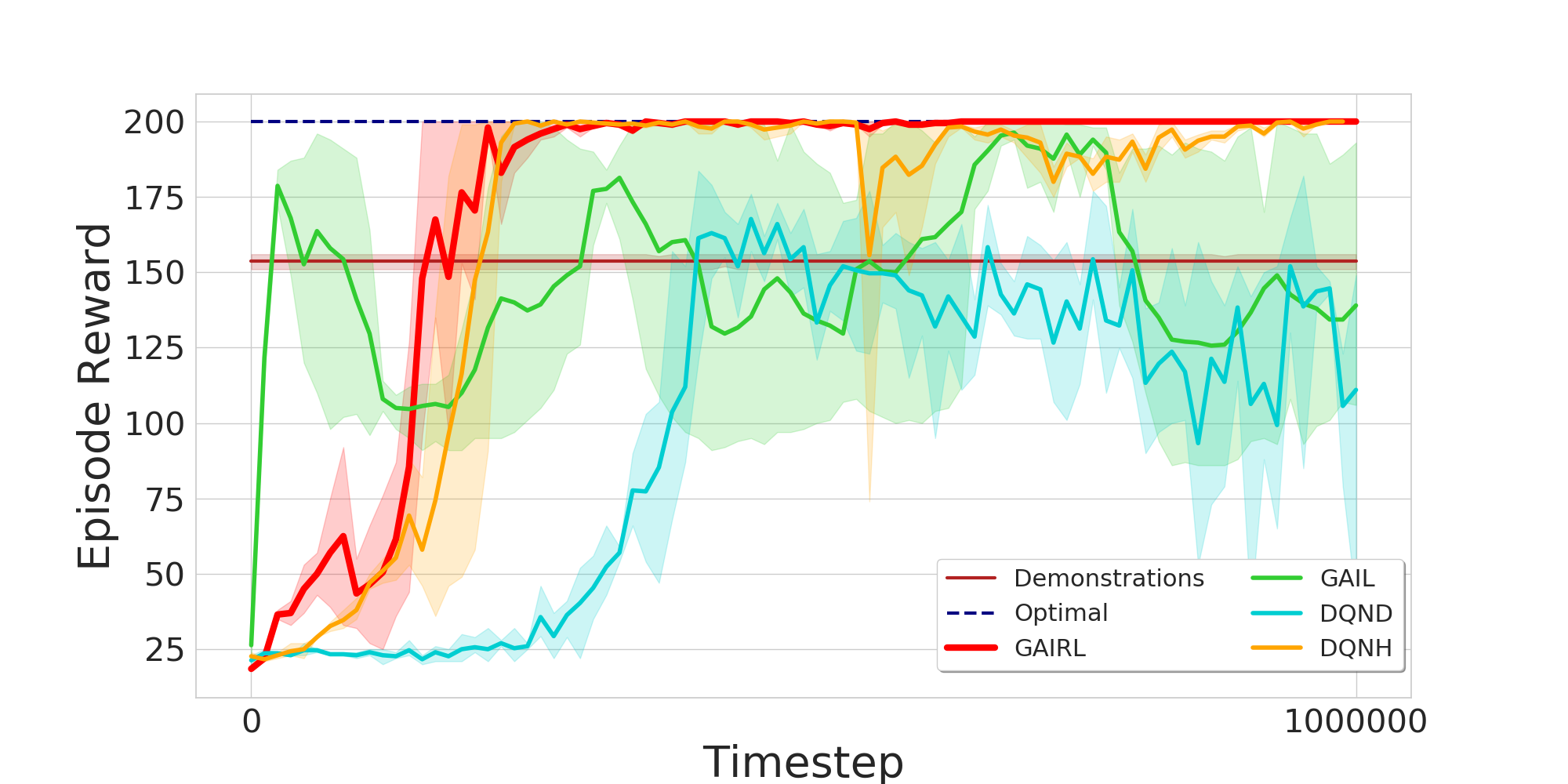}
	\caption{The four agents' learning curves with suboptimal demonstrations 
		in Cart Pole.}
	\label{cartpoleS}
	\vspace{-1mm}
\end{figure}

Figure \ref{cartpoleS} shows four agents' learning curves with suboptimal demonstrations in Cart Pole. 
As shown in Figure \ref{cartpoleS}, even with suboptimal demonstrations, the performance of the GAIRL agent still reaches the optimal performance in the first peak and stabilizes afterwards, 
while the GAIL agent's performance fluctuates dramatically around the performance of demonstrations with a faster learning speed. 
Same to results in Figure \ref{cartpoleO}, while still learning slower, 
the performance of the DQND agent fluctuates around the performance of demonstrations 
and the DQNH agent learns much faster and obtains a stable optimal policy finally. 

\begin{figure}[thpb]
	\centering
	\vspace{-1mm}
	\includegraphics[width=3.6in,height=2in]{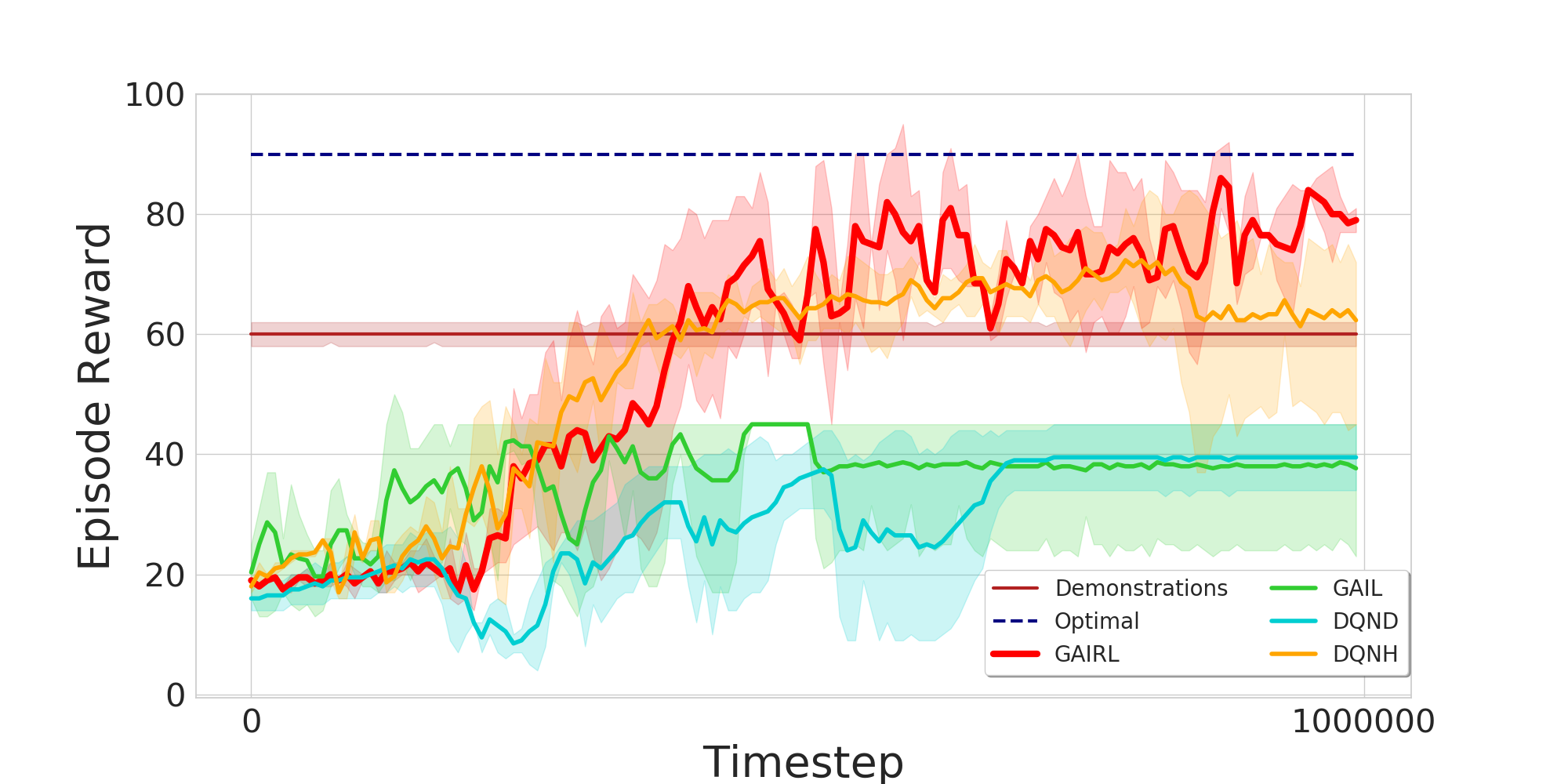}
	\caption{The four agents' learning curves with suboptimal demonstrations in Mountain Car.} 
	\label{mc}
	\vspace{-3mm}
\end{figure}

The four agents' learning with suboptimal demonstrations were also tested in Mountain Car, as shown in Figure \ref{mc}. 
From Figure \ref{mc} we can see that, similar to Figure \ref{cartpoleS}, even with far from optimal demonstrations, the performance of the GAIRL agent can still 
surpass the demonstrations and be close to the optimal one. In contrast, although learning faster, the performance of the GAIL agent is 
far from the performance of demonstrations. 
The DQND agent still learns much slower than the GAIL and GAIRL agent, reaching a similar performance to the GAIL agent finally. However, the DQNH agent learns much faster and its final performance is better than the demonstrations but worse than the GAIRL agent. 

\hfil
\subsubsection{High-dimensional Tasks}

We also tested and compared the performances of the four agents with suboptimal demonstrations in four high-dimensional tasks: Inverted Double Pendulum, Lunar Lander, Hopper and HalfCheetah. 

\begin{figure}[h]
\vspace{-3mm}
	\centering
	\includegraphics[width=3.6in,height=2in]{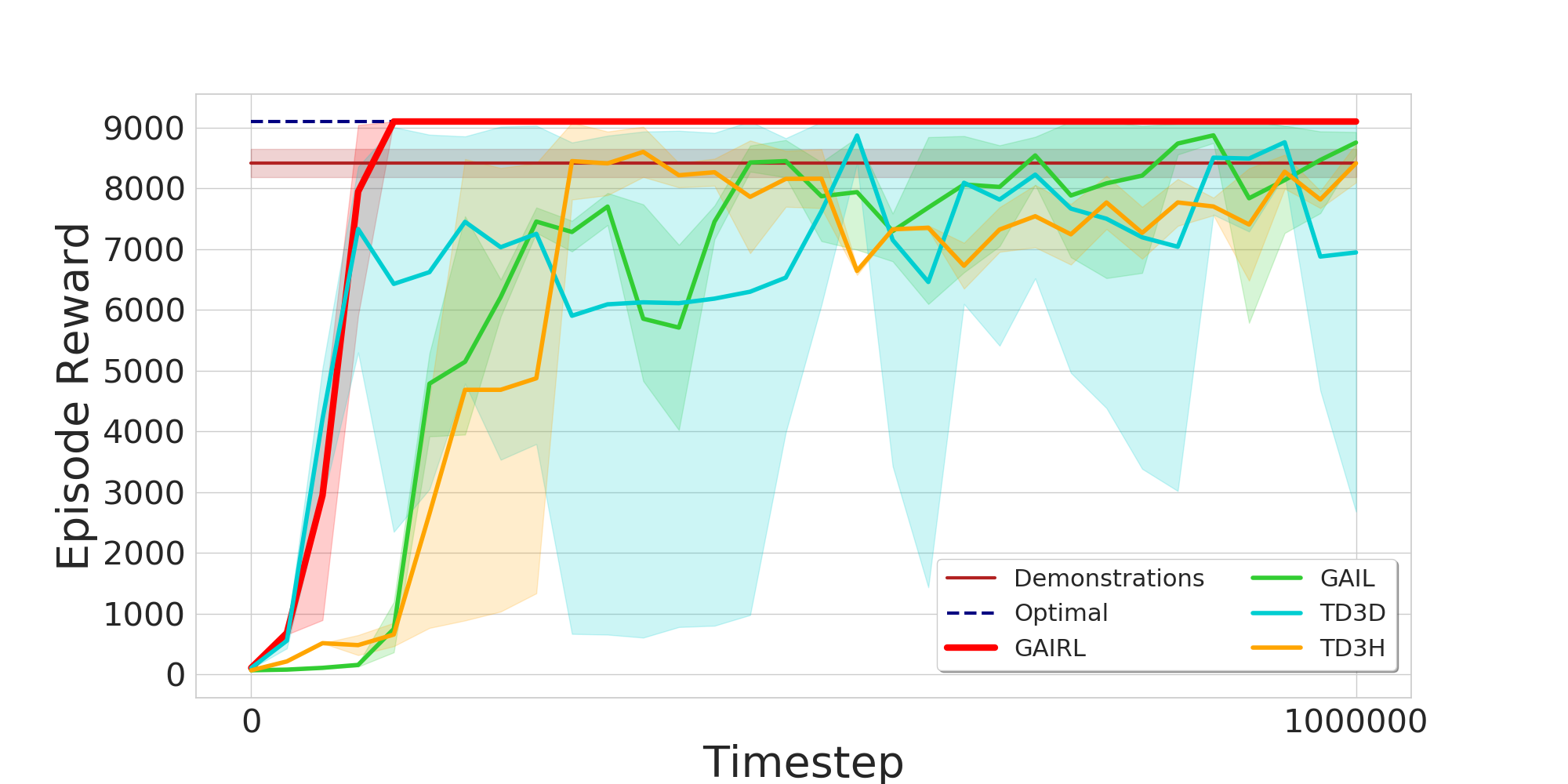}
	\caption{The four agents' learning curves with suboptimal demonstrations 
	in Inverted Double Pendulum.}
	\vspace{-1mm}
	\label{idp}
\end{figure}

Figure \ref{idp} shows four agents' learning curves in Inverted Double Pendulum. Different from results in the low-dimensional tasks, Figure \ref{idp} shows both GAIRL and TD3D agents learn much faster than the GAIL agent, which might be because of the superiority of TD3 used in GAIRL. 
The GAIRL agent achieves an optimal performance in the first peak and stabilizes immediately after that, 
while the performance of GAIL,TD3D and TD3H agents fluctuate around the performance of demonstrations. 

\begin{figure}[h]
	\centering
	\vspace{-3mm}
	\includegraphics[width=3.6in,height=2in]{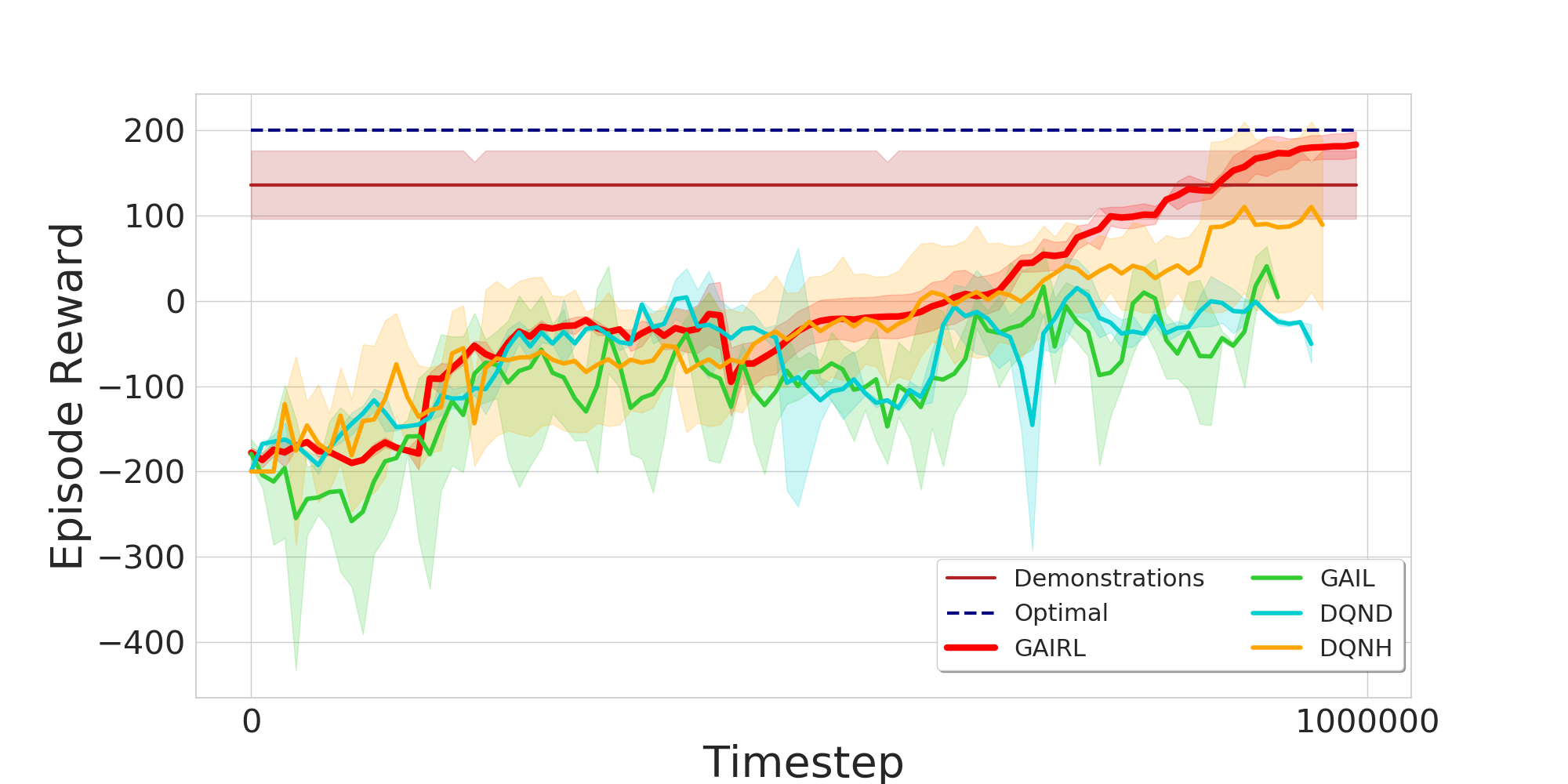}
	\vspace{-2mm}
	\caption{The four agents' learning curves with suboptimal 
	demonstrations 
	in Lunar Lander.}
	\label{lunar}
	\vspace{-2mm}
\end{figure}

Figure \ref{lunar} shows four agents' learning curves in Lunar Lander. 
Different from the Inverted Double Pendulum task, the performance of demonstrations used here is less stable with a relatively high variance. 
From Figure \ref{lunar} we can see that, 
while both the GAIL and DQND agents learn slowly at the beginning, deteriorate in the middle of training 
and reach a performance far worse than demonstrations, the GAIRL agent reaches a close to optimal performance after a long training time finally. 
The DQNH agent's learning is quite similar to that of the GAIRL agent, with a final performance worse than the demonstrations but better than those of the GAIL and DQND agents.

\begin{figure}[thpb]
	\centering
	\vspace{-3mm}
	\includegraphics[width=3.6in,height=2in]{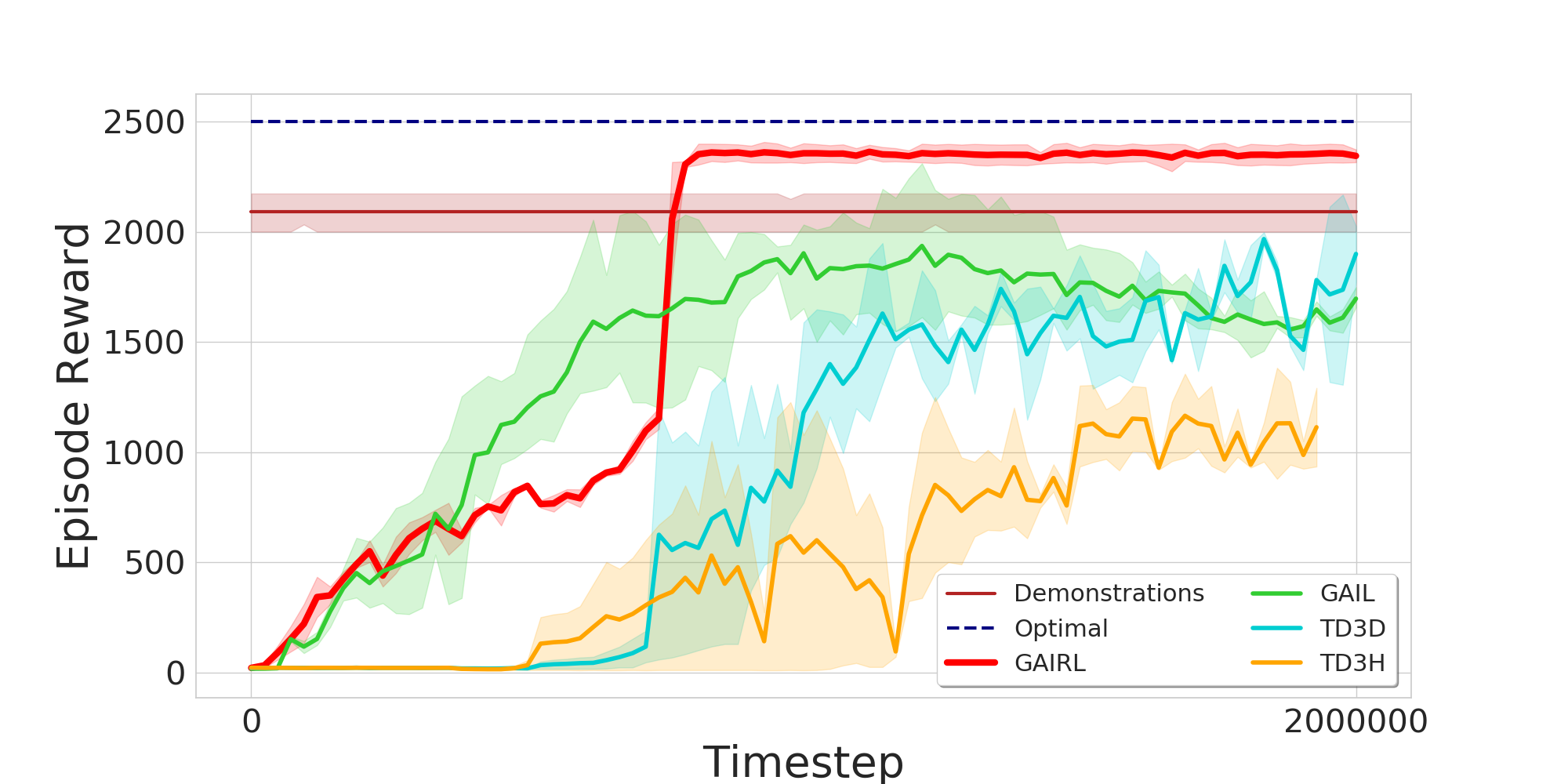}
	\vspace{-2mm}
	\caption{The four agents' learning curves with suboptimal 
	demonstrations 
	in Hopper.}
	\label{hpl}
	\vspace{-2mm}
\end{figure}

Figure \ref{hpl} shows four agents' learning curves in Hopper. 
From Figure \ref{hpl} we can see that, the GAIL agent learns a bit faster than the GAIRL agent but with a final performance worse than the demonstrations, while the GAIRL agent achieves a performance much better than the demonstrations in the first peak after a third of the training time and stabilizes after that. Both TD3D and TD3H agents learn much slower than the GAIL and GAIRL agents, while the final performance of the TD3D agent is similar to that of GAIL agent and the TD3H agent achieves a far worse performance. 

\begin{figure}[thpb]
	\centering
	\vspace{-3mm}
	\includegraphics[width=3.6in,height=2in]{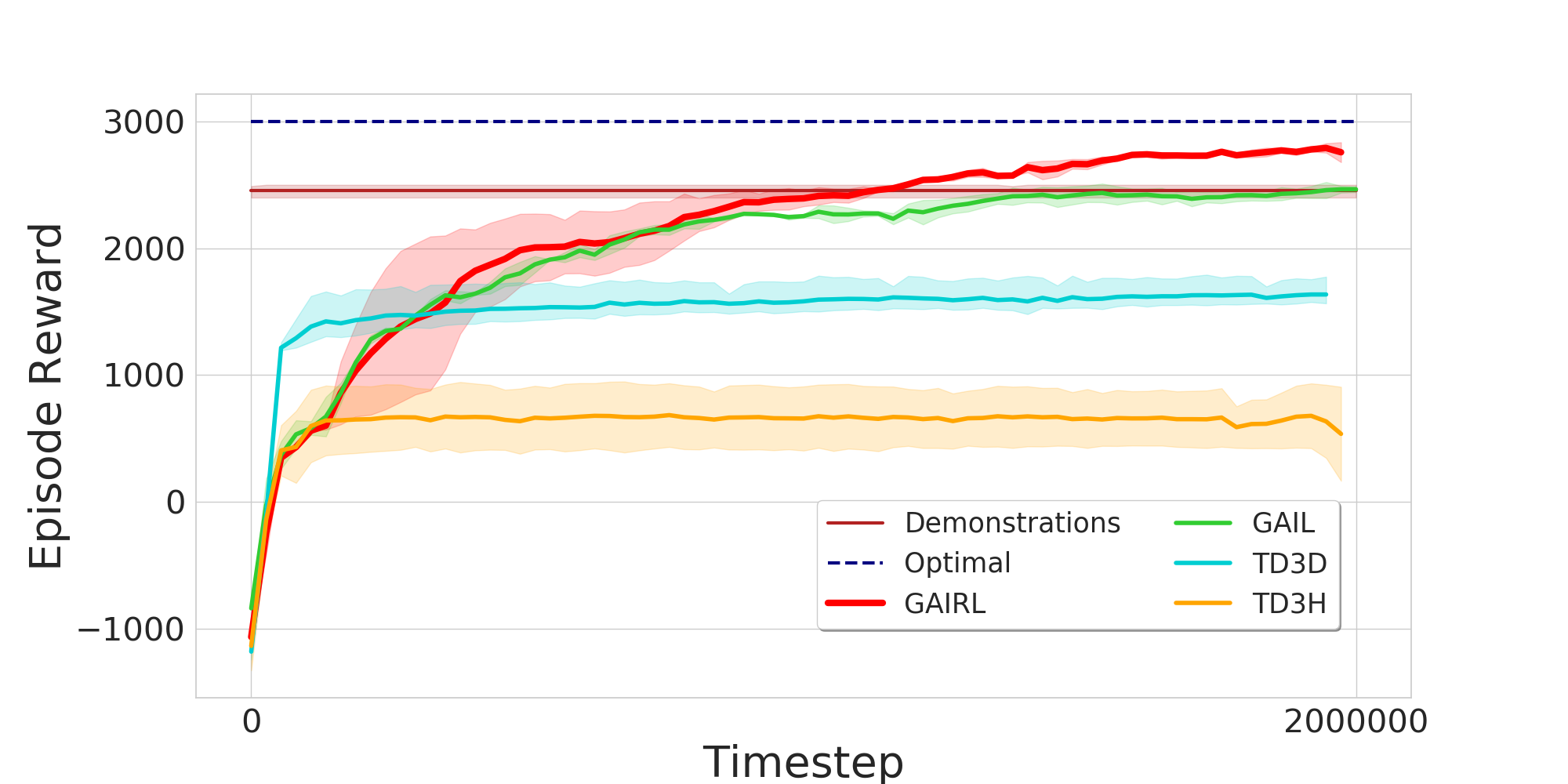}
	\vspace{-2mm}
	\caption{The four agents' learning curves with suboptimal 
	demonstrations 
	in HalfCheetah.}
	\label{hcl}
	\vspace{-2mm}
\end{figure}

Figure \ref{hcl} shows four agents' learning curves in HalfCheetah. 
Different from other tasks, all four agents learn fast in HalfCheetah with the speed of the TD3D agent a bit faster than the other three agents. However, both performances of the TD3H agent and TD3D agent stabilize after reaching the first peak and are far worse than the performance of demonstrations. In contrast, both performances of the GAIL agent and GAIRL agent rise steadily in a similar way.  The GAIRL agent's final performance is close to optimal and better than that of demonstrations, while the GAIL agent reaches a final performance almost the same as that of demonstrations.
\hfil
\subsubsection{Stability Analysis}

\begin{figure}[h]
	\vspace{-3mm}
	\centering
	\includegraphics[width=3in,height=2in]{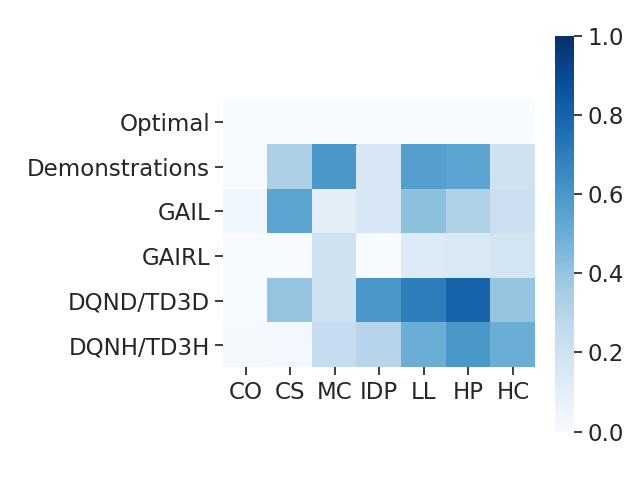}
	\caption{Variance of the four agents' final policies compared to the expert and optimal policy in the six tasks. Note: CO---Cart Pole with Optimal Demonstration, CS---Cart Pole with Suboptimal Demonstration, MC---Mountain Car, IDP---Inverted Double Pendulum, LL---Lunar Lander, HP---Hopper, HC---HalfCheetah.}
	\vspace{-2mm}
	\label{variance}
	\vspace{-1mm}
\end{figure}
We also studied the stability of the final learned policy of the four agents by testing them for 10 times in each task and computing the variance over the 10 performances for each policy. The variances of all policies, including the demonstrated policy and the optimal policy, are shown in Figure \ref{variance}. The optimal policy is always stable while the stability of the demonstrated policy is different in each task.  
Figure \ref{variance} shows that the GAIRL agent always learns a much more stable policy than the demonstrated policy and most of the time almost as stable as the optimal one, while the GAIL agent's policy is only slightly more stable than the demonstrated policy and even a bit less stable in Cart Pole with both optimal and suboptimal demonstrations. 
In addition, the policy of DQND/TD3D and DQNH/TD3H agents become generally less stable as the task's difficulty increases (from left to right in Figure \ref{variance}), while the stability of the GAIRL agent keeps at a similar and much higher level than both of them. This suggests the demonstrations and human evaluative feedback might have a complementary effect. Demonstrations provide a high-level initialization of the human's overall reward function, while human evaluative feedback like preferences can explore specific, fine-grained aspects of it \cite{biyik2020learning}. 

In summary, our results suggest that for both low-dimensional and high-dimensional tasks with both optimal and suboptimal demonstrations, a GAIRL agent can learn a more stable policy with optimal or close to optimal performance, while the performance of the GAIL agent is upper bounded by the performance of demonstrations 
or even worse than it. In addition, our results indicate the superiority of GAIRL over GAIL might be because of the complementary effect of the demonstrations and human evaluative feedback.

\section{Conclusion}

In this paper, to address the limit of GAIL that it seldom surpasses the performance of demonstrations, we propose a model-free framework --- GAN-Based Interactive Reinforcement Learning (GAIRL) by combining the advantages of GAIL and interactive RL from human evaluative feedback. We tested GAIRL in six physics-based control tasks 
from the classic RL literature. Our results suggest that with both optimal and suboptimal demonstrations, the GAIRL agent can learn a more stable policy with optimal or close to optimal performance, while the performance of the GAIL agent is upper bounded by the the performance of demonstrations 
or even worse than it. In addition, our results indicate the superiority of GAIRL over GAIL might be because human evaluative feedback is complementary to demonstrations. 

%
\ifCLASSOPTIONcaptionsoff
  \newpage
\fi

%
\bibliographystyle{IEEEtran}
\bibliography{IEEEabrv,GAIRL}
\newpage
\begin{IEEEbiography}[{\includegraphics[width=1in,height=1.25in,clip,keepaspectratio]{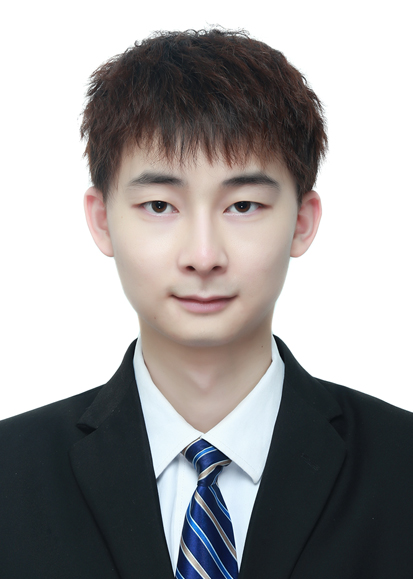}}]{Jie Huang} received a Bachelor degree in electronic information science and technology from School of Information Science and Engineering, Ocean University of China in 2019. He is currently pursuing  the master's degree with the School of Information Science and Engineering, Ocean University of China, Qingdao, China. His current research interests include reinforcement learning, human agent/robot interaction, imitation learning and robotics.
\end{IEEEbiography}
\begin{IEEEbiography}[{\includegraphics[width=1in,height=1.25in,clip,keepaspectratio]{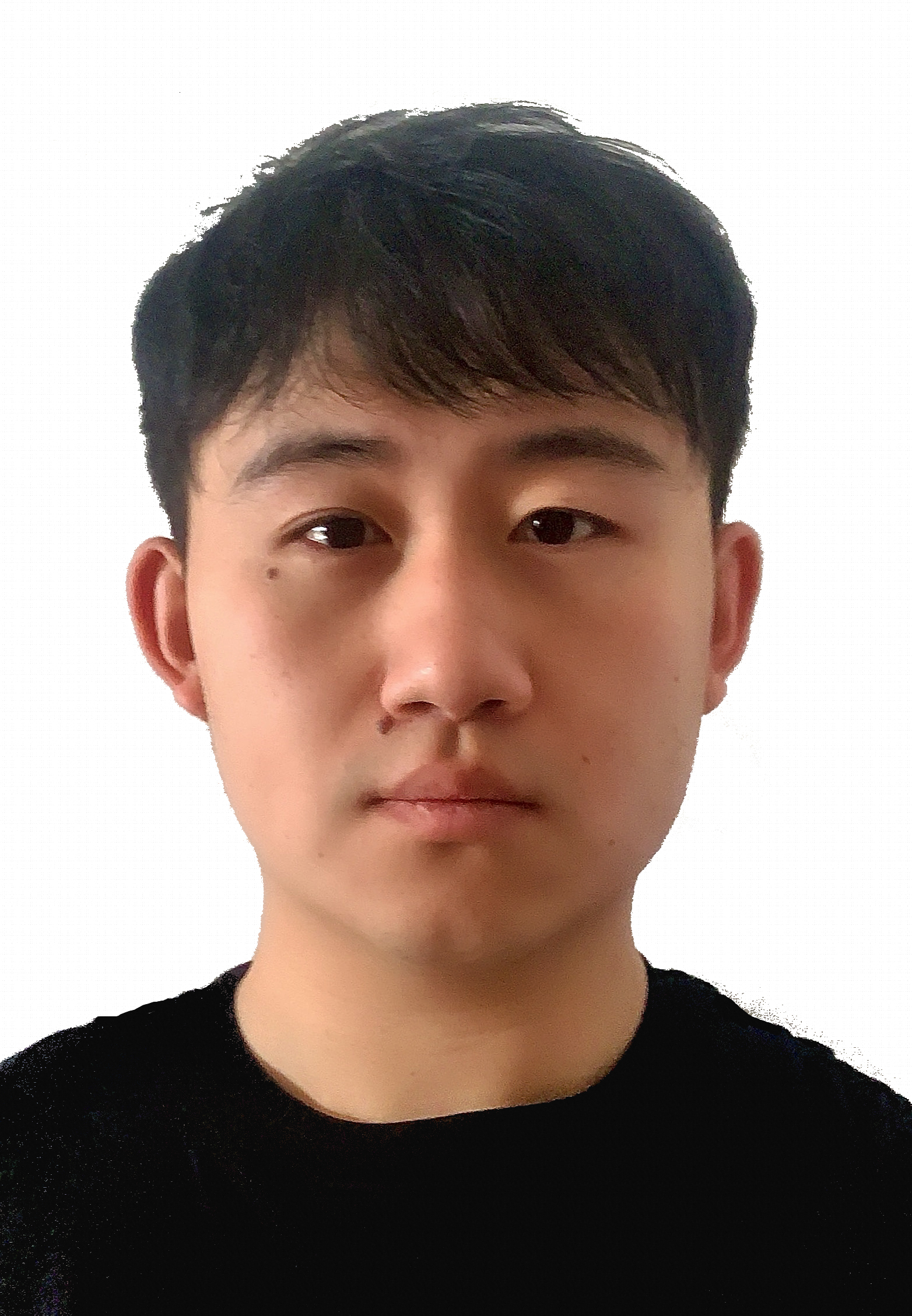}}]{Rongshun Juan} received a Bachelor degree in electronic information science and technology from School of Information Science and Engineering, Ocean University of China in 2019. He is currently pursuing  the master's degree with the School of Information Science and Engineering, Ocean University of China, Qingdao, China. His current research interests include reinforcement learning, human agent/robot interaction, transfer learning and robotics.
\end{IEEEbiography}
\begin{IEEEbiography}[{\includegraphics[width=1in,height=1.25in,clip,keepaspectratio]{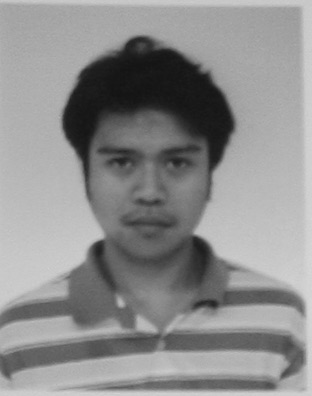}}]{Randy Gomez}
received M.Eng.Sci. in Electrical
Engineering at the University of New South Wales (UNSW), Australia in
2002. He obtained his Ph.D. in 2006 from the Graduate
School of Information Science, Nara Institute of
Science and Technology, Japan. He was a researcher at Kyoto University
until 2012 under the auspices of the Japan Society for the Promotion of Science (JSPS) research fellowship.
Currently, he is a senior scientist at Honda Research Institute Japan. His research interests include
robust speech recognition, acoustic modeling and adaptation, multimodal interaction and robotics.
\end{IEEEbiography}
\begin{IEEEbiography}[{\includegraphics[width=1in,height=1.25in,clip,keepaspectratio]{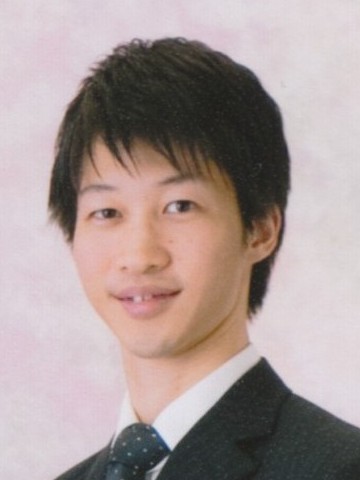}}]{Keisuke Nakamura}
received a B.E. in 2007 from the Department of Control and System
Engineering, Tokyo Institute of Technology, Japan.  He also studied in
the Department of Electronic and Electrical Engineering at the
University of Strathclyde, United Kingdom from 2007 to 2008. He
received an M.E. from the Department of Mechanical and Control
Engineering at Tokyo Institute of Technology in 2010, and Ph.D. in
Informatics in 2013 from Kyoto University, Japan.  He is currently a
senior scientist for Honda Research Institute Japan, Co., Ltd. His
research interests are in the field of robotics, control systems, and
signal processing. He is currently a member of IEEE.
\end{IEEEbiography}
\begin{IEEEbiography}[{\includegraphics[width=1in,height=1.25in,clip,keepaspectratio]{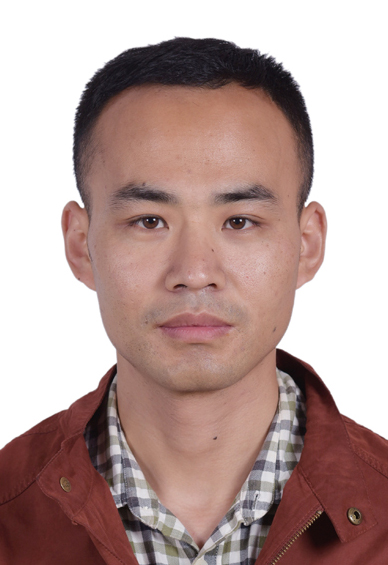}}]{Qixin Sha} received the B.S. degree in communication engineering from Ocean University of China in 2007, and the M.S. degree in communication and information systems from Ocean University of China in 2010. He worked at Qingdao Bailing Technology Co., Ltd. and Alcatel-Lucent Qingdao R \& D Center from 2010 to 2014 as a software engineer. He is currently working as an experimenter in the Department of Electronic Engineering, Ocean University of China. His research interests include the design and development of architecture, decision-making and software system in underwater vehicle.
\end{IEEEbiography}
\begin{IEEEbiography}[{\includegraphics[width=1in,height=1.25in,clip,keepaspectratio]{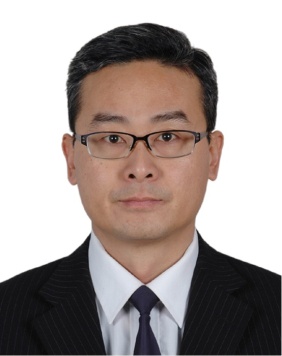}}]{Bo He}
received a PhD degree from Harbin Institute of Technology in 1999. He was a researcher in Nanyang Technological University from 2000 to 2002. He is currently a full professor in Ocean University of China. His research interests include SLAM, machine learning and robotics. He is currently a member of IEEE.
\end{IEEEbiography}
\newpage
\begin{IEEEbiography}[{\includegraphics[width=1in,height=1.25in,clip,keepaspectratio]{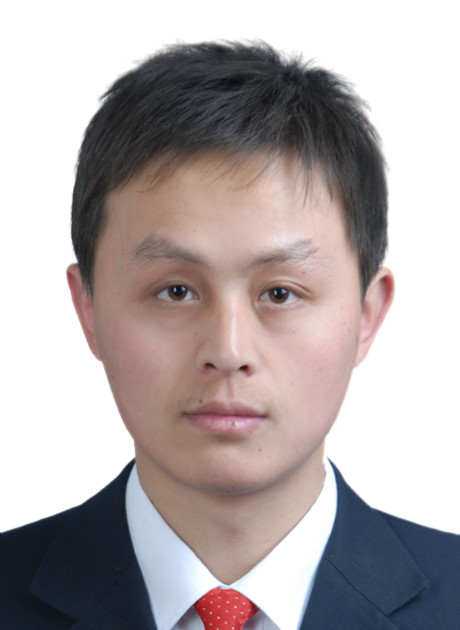}}]{Guangliang Li} received a Bachelor and M.Sc. degree from School of Control Science and Engineering in Shandong University in 2008 and 2011 respectively. In 2016, he received a Ph.D. degree in Computer Science from the University of Amsterdam, 
The Netherlands. He was a visiting researcher in Delft University of Technology, The Netherlands, and a research intern in Honda Research Institute Japan, Co., Ltd. Japan. He is currently a lecturer at Ocean University of China. His research interests include reinforcement learning, human agent/robot interaction and robotics. He is a member of IEEE.
\end{IEEEbiography}

\end{document}